\documentclass[letterpaper, 10 pt, conference]{ieeeconf}  % Comment this line out if you need a4paper

\IEEEoverridecommandlockouts                              % This command is only needed if 
\usepackage{graphics} % for pdf, bitmapped graphics files
\usepackage{epsfig} % for postscript graphics files
\usepackage{mathptmx} % assumes new font selection scheme installed
\usepackage{times} % assumes new font selection scheme installed
\usepackage{amsmath} % assumes amsmath package installed
\usepackage{amssymb}  % assumes amsmath package installed

\usepackage{hyperref}
\usepackage{stix}
\usepackage{mleftright,mathtools}

\usepackage[dvipsnames]{xcolor}
\def\etal{et~al.~}			  % and others, and co-workers
\usepackage{amsmath}
\DeclareMathOperator*{\argmax}{arg\,max}

\makeatletter
\def\thickhline{%
  \noalign{\ifnum0=`}\fi\hrule \@height \thickarrayrulewidth \futurelet
   \reserved@a\@xthickhline}
\def\@xthickhline{\ifx\reserved@a\thickhline
               \vskip\doublerulesep
               \vskip-\thickarrayrulewidth
             \fi
      \ifnum0=`{\fi}}
\makeatother

\newlength{\thickarrayrulewidth}
\definecolor{highlight}{HTML}{70AD47}
\definecolor{wctseng}{HTML}{59D55B}
\definecolor{yclin}{HTML}{ED7D31}
\definecolor{hank}{HTML}{7030A0}

\title{\LARGE \bf
CLA-NeRF: Category-Level Articulated Neural Radiance Field
}

\author{Wei-Cheng Tseng$^{1}$, Hung-Ju Liao$^{1}$, Lin Yen-Chen$^{2}$ and Min Sun$^{1,3}$% <-this % stops a space
\thanks{$^{1}$National Tsing Hua University}%
\thanks{$^{2}$Massachusetts Institute of Technology}%
\thanks{$^{3}$Appier Inc.}
}

\begin{document}

\maketitle
\thispagestyle{empty}
\pagestyle{empty}

\begin{figure*}[ht]
    \centering
    \includegraphics[width=1\textwidth]{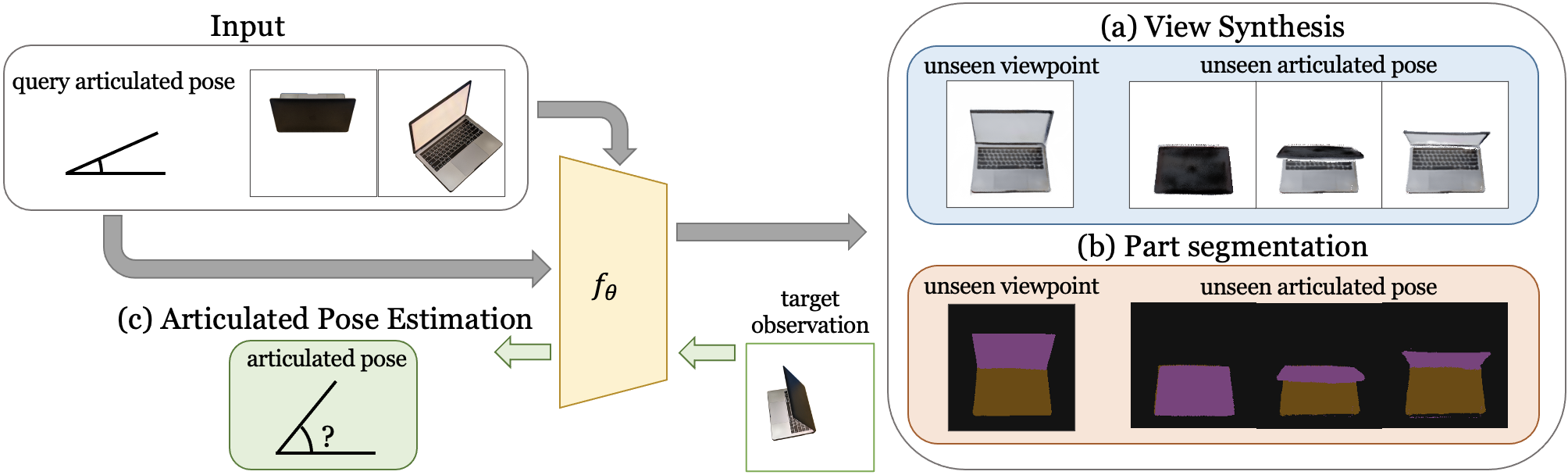}
    \vspace{-1.\baselineskip}
    \caption{
    We present a framework that takes a few visual observations and corresponding camera poses as input; then, we can perform (a) view synthesis and (b) part segmentation from unseen viewpoints and articulated poses. Moreover, (c) the articulated pose can be estimated via inversely optimizing the 3D deformation through our framework to match the target visual observation.}
    \vspace{-1.\baselineskip}
    \label{fig:teaser}
\end{figure*}

\begin{abstract}
We propose CLA-NeRF -- a Category-Level Articulated Neural Radiance Field that can perform view synthesis, part segmentation, and articulated pose estimation. CLA-NeRF is trained at the object category level using no CAD models and no depth, but a set of RGB images with ground truth camera poses and part segments. During inference, it only takes a few RGB views (i.e., few-shot) of an unseen 3D object instance within the known category to infer the object part segmentation and the neural radiance field. Given an articulated pose as input, CLA-NeRF can perform articulation-aware volume rendering to generate the corresponding RGB image at any camera pose. Moreover, the articulated pose of an object can be estimated via inverse rendering. In our experiments, we evaluate the framework across five categories on both synthetic and real-world data. In all cases, our method shows realistic deformation results and accurate articulated pose estimation. We believe that both few-shot articulated object rendering and articulated pose estimation open doors for robots to perceive and interact with unseen articulated objects. 
% Please see \url{https://bit.ly/3iWoque} for qualitative results.
\end{abstract}
\section{Introduction}
Our living environment is full of articulated objects: objects composed of more than one rigid parts (links) connected by joints allowing rotational or translational motion, such as doors, refrigerators, scissors, and laptops. Endowing robots with the ability to perceive and interact with these objects requires a detailed understanding of the objects' part-level poses, 3D shape, and materials. 
% Add a sentence on why this problem is hard.
Prior works~\cite{desingh2019factored,BMVC2015_181} on estimating these properties of articulated objects often assume the object's CAD model and thus cannot generalize to objects unseen during training.

To address this limitation, several recent works have explored category-level representations for articulated objects. These representations do not assume CAD models during testing and therefore can achieve intra-category generalization. For instance, ANCSH~\cite{cvpr20_ANCSH}, designed specifically for articulated object pose estimation, uses the 3D coordinates in the canonical frame as the representation where the canonical frame is determined by authors who manually align the center and orientation of different CAD models. A-SDF~\cite{arxiv21_asdf}, focusing on articulated object shape reconstruction, uses deep implicit signed distance function~\cite{cvpr19_deepsdf} as the representation and factors the latent space into shape codes and articulation angles.
Although these representations have shown impressive results, both of them are limited by the requirement of access to ground truth 3D geometry during training, which is costly to scale up for articulated objects~\cite{3DIndoorObjects}. Furthermore, during testing, both works require depth images as inputs. This poses additional requirements on hardware and may not work on articulated objects that are thin or highly reflective, e.g., scissors.

% \begin{figure*}
%     \centering
%     \includegraphics[width=1\textwidth]{figures/teaser_v7.png}
%     \caption{
%     We present a framework that takes a few visual observations and corresponding camera poses as input; then, we can perform (a) view synthesis and (b) part segmentation from unseen viewpoints and articulated poses. Moreover, (c) the articulated pose can be estimated via inversely optimizing the 3D deformation through our framework to match the target visual observation.
%     }
%     \label{fig:teaser}
% \end{figure*}

In this work, we seek to relax these requirements and build a category-level representation for articulated objects that doesn't require 3D CAD models or depth sensing during both training and testing --- only using RGB images with camera poses and part segmentation labels for training, and RGB images alone for testing. To this end, we introduce CLA-NeRF, a Category-Level Articulated NeRF representation that supports multiple downstream tasks including novel view synthesis, part segmentation, and articulated pose estimation. Our representation is based on Neural Radiance Fields (NeRF~\cite{eccv20_nerf}), a method that has shown impressive performance on novel view synthesis of a specific scene by encoding volumetric density and color through a neural network. As NeRF typically requires a lengthy optimization process for each scene independently, we follow recent works~\cite{cvpr21_pixelnerf,wang2021ibrnet} to directly predict NeRFs from one or several RGB images in a feed-forward manner. However, simply doing so cannot capture articulated objects' part attributes (e.g., part poses and segmentation) and joint attributes (e.g., joint axis). We, therefore, propose to explicitly model the object articulation by predicting a part segmentation field in addition to the volumetric density and color. Joint attributes can then be inferred by performing line-fitting on the part segmentation field. 
In the experiments, we focus on modeling objects with revolute joints that cause 1D rotational motion (e.g., eyeglasses). We show that CLA-NeRF can render the object and its part segmentation map at unseen articulated poses by performing articulation-aware volume rendering. Additionally, it can perform category-level articulated pose estimation with RGB inputs by minimizing the residual between the rendered and observed pixels. We note that these tasks are not possible with existing NeRF formulations~\cite{eccv20_nerf,cvpr21_pixelnerf,wang2021ibrnet} which explicitly model the camera poses but don't consider the object articulation. To the best of our knowledge, our work is the first to model general articulated objects with neural radiance fields. 
We summarize our primary contributions as follows, and more information are provided in our project website\footnote{https://weichengtseng.github.io/project\_website/icra22/index.html}:
\begin{itemize}
    \item We propose CLA-NeRF, a differentiable representation for articulated objects that explicitly models the part and joint attributes. The proposed representation disentangles camera pose, part pose, part segmentation, and joint attributes, allowing us to independently control each property during rendering.
    \item We show that the proposed representation can perform category-level articulated pose estimation through analysis-by-synthesis with only RGB inputs. To the best of our knowledge, existing works for this task all require depth inputs~\cite{cvpr20_ANCSH,weng2021captra,arxiv20_screwnet,corl20_learn}.
    % Maybe add one more contribution?
\end{itemize}

\section{Related Works}
\vspace{-0.3\baselineskip}
\subsection{Articulated 3D Shape Representations}
% \paragraph{Articulated 3D Shape Representations}
Meshes and rigging techniques~\cite{skinningcourse:2014} are widely used to model the shape and deformation of articulated objects. Leveraging the abundant prior knowledge of human bodies, efficient techniques~\cite{SMPL2015,Zheng2019DeepHuman,bhatnagar2019mgn,peng2021neural,kocabas2019vibe,hmrKanazawa17,zhang2020phosa,omran2018nbf,aaai21_tseng,meta-cpr}1 have been developed to model the deformation of a wide variety of body shapes. However, creating watertight meshes and rigs remains a labor-intensive process for general articulated objects whose part and joint attributes are less constrained. For the robotics community, it is very costly, if not impossible, to hire specially trained experts to model all sorts of articulated objects exist in our daily life. 
Recently, NASA~\cite{deng2019neural} proposes to represent articulated shapes with a neural indicator function that successfully circumvents the complexity of meshes and the issue of water-tightness. A-SDF~\cite{arxiv21_asdf} uses neural networks to encode signed distance function for articulated shape modeling. It's trained on multiple instances of the same category and learns a disentangled latent space that allows it to synthesize novel shapes at unseen articulated poses. However, both of them require ground truth 3D models for training and thus still suffer from the scalability issue.
Concurrently with our work, NARF~\cite{2021narf} also proposes to explicitly consider articulation within NeRF and show impressive results on view synthesis of human bodies. Compared to NARF, our method differs in two aspects. First, our method focuses on general articulated objects and thus doesn't assume known joint attributes (e.g., root joint's pose, bone length)  during test time. Instead, we infer them from the predicted segmentation field and further show results on articulated pose estimation. Second, our method uses RGB images and part segmentation labels as supervision, while NARF uses RGB images and joint attributes. We believe both works complement each other and further supports the possibility that explicitly considering articulation within NeRF can lead to better generalization.
% \vspace{-0.5\baselineskip}
%
\vspace{-0.3\baselineskip}
\subsection{Articulated Object Pose Estimation}
% \paragraph{Articulated Object Pose Estimation}
Most existing approaches for articulated object pose estimation requires instance-level information. They either assume the articulated object's exact CAD model~\cite{desingh2019factored,BMVC2015_181} or need to generate the object's motion through deliberate interaction before inference~\cite{katz2008manipulating,katz2013interactive,martin2014online,martin2016integrated,hausman2015active,arxiv15_visual}. Both directions require the robot to learn about each object from scratch, no matter how similar the object is to those it has previously experienced. To address this issue, recent works have proposed to predict canonicalized object coordinates~\cite{wang2019normalized} for category-level articulated object pose estimation~\cite{cvpr20_ANCSH,weng2021captra}. However, such representation is designed specifically for articulated pose estimation and can't perform other tasks such as shape reconstruction or view synthesis. Additionally, it requires articulated objects' ground truth 3D geometries for training and depth images for testing. 
As for inferring articulated pose from visual data, \cite{corl20_learn} proposed to use a mixture density network that consumes an RGB-D image to predict the probability of the joint attribute and articulated pose. ScrewNet\cite{arxiv20_screwnet} takes multiple depth images with different articulated poses and the same camera pose as input to predict joint attribute and articulated pose. \cite{iros20_learn} extended \cite{arxiv11_prob} by including reasoning about the applied actions along with the observed motion of the object while estimating its kinematic structure.
Different from these works, we focus on building a category-level representation that only requires 2D supervision. Also, we demonstrate results on view synthesis besides articulated pose estimation.
%
% \vspace{-0.3\baselineskip}
\begin{figure*}[t]
    \centering
    \includegraphics[width=0.94\textwidth]{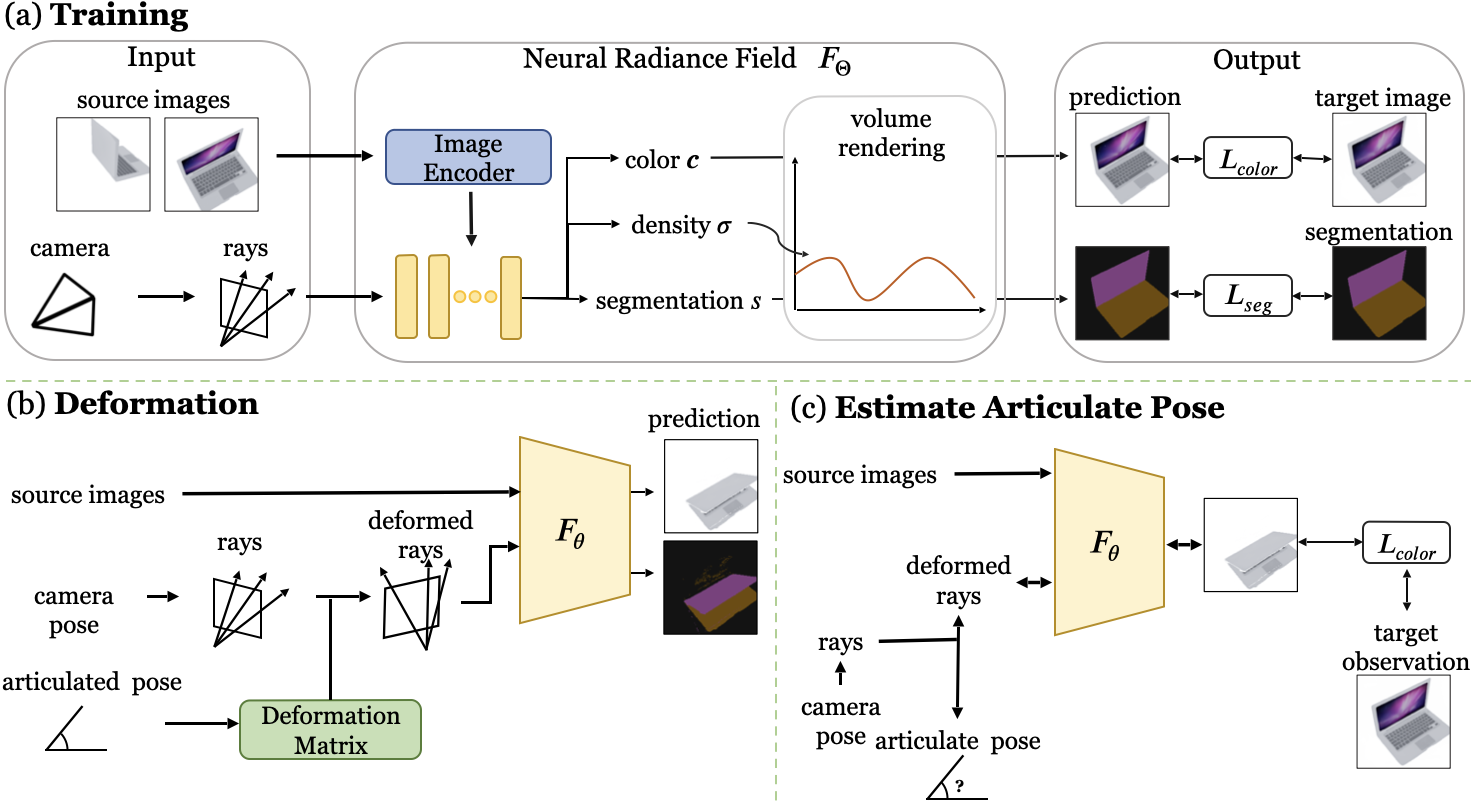}
    \vspace{-1\baselineskip}
    \caption{The overview of our framework. 
    (a) Our framework retrieves features from two instance as the condition of NeRF model and predicts color $\mathbf{c}$, density $\sigma$ and segmentation $\mathbf{s}$. The volume rendering is applied to predict rendered results.
    (b) We calculate the deformation matrix based on the articulated pose. Then, we deform the sampled rays with the deformation matrix. Finally, the deformed visual image is rendered using our learned framework. 
    (c) The articulated pose is estimated via inversely minimizing $\mathcal{L}_{\text{color}}$.}
    \label{fig:overview}
    \vspace{-1.5\baselineskip}
\end{figure*}
\vspace{-0.3\baselineskip}
\subsection{Preliminaries: NeRF}\label{sec:background}
NeRF learns to synthesize novel views associated with unseen camera poses given a collection of RGB images with known camera poses.
Specifically, NeRF represents a scene as a volumetric field of density $\sigma$ and RGB color $\mathbf{c}$. The density models the shape of the scene and the color models the view-dependent appearance of occupied regions of the scene, both of which lie within a bounded 3D volume.
A multilayer perceptron (MLP) parameterized by the weights $\Theta$ is used to predict the density $\sigma$ and RGB color $\mathbf{c}$ of each point by taking its 3D position $\mathbf{x} = (x, y, z)$ and unit-norm viewing direction $\mathbf{d}$ as input, where $(\sigma, \mathbf{c}) \leftarrow F_{\Theta}(\gamma(\mathbf{x}), \mathbf{d})$ and $\gamma(\cdot)$ is a high-frequency positional encoding \cite{neurips17_attention}.
To render a pixel, NeRF emits a camera ray $\mathbf{r}(t) = \mathbf{o} + t\mathbf{d}$ from the camera center $\mathbf{o}$ along the direction $\mathbf{d}$ passing through that pixel on the image plane. Along the ray, $K$ points $\{\mathbf{x}_k = \mathbf{r}(t_k)\}_{k=1}^K$ are sampled for use as input to the MLP which outputs a set of densities and colors $\{\sigma_k, \mathbf{c}_k\}_{k=1}^K$. These values are then used to estimate the color $\hat{\textbf{C}}(r)$ of that pixel following volume rendering~\cite{kajiya84} approximated with numerical quadrature~\cite{max95}:
\begin{equation} \label{equ:volume_rendering}
\begin{split}
    \hat{\textbf{C}}(\mathbf{r}) = \sum_{k=1}^{K} \hat{T}_k (1- \exp(-\sigma_k (t_{k+1} - t_k)))\, \textbf{c}_k, \\ \text{with} \quad \hat{T}_k = \text{exp} (-\sum_{k' < k} \sigma_{k'} (t_{k'+1} - t_{k'}))
\end{split}
\end{equation}
where $\hat{T}_k$ can be interpreted as the probability of the ray successfully transmits to point $\mathbf{r}(t_k)$. 
NeRF is then trained to minimize a photometric loss $\mathcal{L} = \sum_{\mathbf{r} \in \mathbf{R}} || \hat{\mathbf{C}}(\mathbf{r}) - \mathbf{C}(\mathbf{r})||_2^2$,
using some sampled set of rays $\mathbf{r} \in \mathbf{R}$ where $\mathbf{C}(\mathbf{r})$ is the observed RGB value of the pixel corresponding to ray $\mathbf{r}$ in some image. 
To improve rendering efficiency one may train two MLPs: one ``coarse'' and one ``fine'', where the coarse model serves to bias the samples that are used for the fine model.
For more details, we refer readers to Mildenhall \etal~\cite{eccv20_nerf}. 

While NeRF originally needs to optimize the representation for every scene independently, several recent works~\cite{cvpr21_pixelnerf,wang2021ibrnet,icml21_sharf} on category-level NeRF have been proposed to directly predict a NeRF conditioned on one or few input images. 
\vspace{-0.5\baselineskip}

\section{Method}
% First, talk about why existing NeRF can't handle articulated objects
Although NeRF has shown impressive results on modeling the appearance of static objects, its formulation only allows control over the camera poses during rendering. Therefore, it cannot render a deformable articulated object (e.g., laptop) at different articulated poses (e.g., closing vs. opening) because it has more than 6 degree of freedom (DoF).
% Then, talk about why CLA-NeRF can fix these issues.
CLA-NeRF is designed to tackle these issues. Instead of simply predicting colors $\mathbf{c}$ and densities $\sigma$ for each 3D location, we propose to additionally estimate part segmentation $\mathbf{s}$. Instead of only controlling the camera poses during rendering, our formulation allows user to input articulated poses. And instead of casting rays solely based on camera poses, we also transform camera rays based on query articulated poses, predicted part segmentation, and inferred joint attributes during volume rendering. These modifications together allow CLA-NeRF to render articulated objects at unseen articulated poses.

\vspace{-0.5\baselineskip}

\subsection{Category-Level Semantic NeRF} \label{sec:part_segmentation}

Here we first describe how we extend NeRF to predict part segmentation. For each 3D location $\mathbf{x}$ and viewing direction $\mathbf{d}$, we add another linear layer on top of NeRF's MLP backbone to predict part segmentation: 
\begin{equation}
    (\sigma, \textbf{c}, \textbf{s}) = F_{\Theta}\Big(\gamma(\textbf{x}), \textbf{d}\Big)
\end{equation}
where $\mathbf{s}$ is the segmentation logits with $P+1$ dimension ($P$ parts and background).

With the volumetric field of predicted part segmentation, we can predict which part a pixel belongs to following the procedure we used to approximate the volume rendering of RGB:
\begin{equation}
\begin{split}
    \hat{\textbf{S}}(\mathbf{r}) = \sum_{k=1}^{K} \hat{T}_k (1- \exp(-\sigma_k (t_{k+1} - t_k))) \textbf{s}_k, \\ \text{with} \quad \hat{T}_k = \text{exp} (-\sum_{k' < k} \sigma_{k'} (t_{k'+1} - t_{k'}))
\end{split}
\end{equation}
where $\textbf{s}_k$ is the predicted part segmentation of sampled point $\mathbf{r}(t_k)$.
% \textbf{Training.} \label{sec:training}
The new model can then be trained with both color loss $\mathcal{L}_{\text{color}}$ and segmentation loss $\mathcal{L}_{\text{seg}}$:
\begin{equation}
    \mathcal{L}_{\text{color}} = \sum_{\mathbf{r} \in \mathbf{R}}   \left[
    || \hat{\mathbf{C}}_c(\mathbf{r}) - \mathbf{C}(\mathbf{r}) ||_2^2 + || \hat{\mathbf{C}}_f(\mathbf{r}) - \mathbf{C}(\mathbf{r}) ||_2^2 \right]
\end{equation}
\begin{equation}
    \mathcal{L}_{\text{seg}} = -\sum_{\mathbf{r} \in \mathbf{R}} 
    \left[
    \sum_{i=1}^{P+1} p^i(\mathbf{r}) \text{log}\, \hat{p}_{c}^{i}(\mathbf{r}) + p^{i}(\mathbf{r}) \text{log}\, \hat{p}_{f}^{i}(\mathbf{r})
    \right]
\end{equation}
where $\hat{p}^i(\mathbf{r}) = \frac{\text{exp}(\hat{\mathbf{S}}^i(\mathbf{r}))}{\sum_{j=1}^P\text{exp}(\hat{\mathbf{S}}^j(\mathbf{r}))}$

Here, $\mathbf{C}(\mathbf{r})$, $\hat{\mathbf{C}}_c (\mathbf{r})$ and $\hat{\mathbf{C}}_f(\mathbf{r})$ are the ground truth color, color predicted by the coarse network, and color predicted by the fine network for ray $\mathbf{r}$, respectively. In the segmentation loss $\mathcal{L}_{\text{seg}}$, $p^i$ is the ground truth probability of part $i$, while $\hat{p}_c^i$ and $\hat{p}_f^i$ represent the probability predicted by the coarse and fine network for ray $\mathbf{r}$. In summary, the color loss $\mathcal{L}_{\text{color}}$ is the L2 distance between ground truth color and the color predicted by both coarse and fine networks, and the segmentation loss $\mathcal{L}_{\text{seg}}$ is a multi-class cross-entropy loss that encourages the rendered semantic labels to be consistent with the provided labels. A coefficient $\lambda$ is used to modulate these two losses during training:  $\mathcal{L}_{\text{total}} = \mathcal{L}_{\text{color}} + \lambda \cdot \mathcal{L}_{\text{seg}}$.

We note that the current formulation still requires lengthy optimization for each articulated object and does not share knowledge between different objects. 
To make our method generalize to objects within the same category, we customize the framework of previous works~\cite{cvpr21_pixelnerf,wang2021ibrnet} to directly predict the proposed semantic NeRF given one or a few input images of the articulated object. For brevity, we explain the framework with a single input image $I$. First, we extract the image feature with an image encoder $E$ to form a feature map $W = E(\mathbf{I})$. Then, we project each sampled 3D point $\mathbf{x}$ to the input image plane and get the projected coordinate $\pi(\mathbf{x})$. Finally, we augment the input to NeRF $F_{\Theta}$ with the associated feature $W(\pi(\textbf{x}))$, resulting in the following formulation:
\begin{equation}
    (\sigma, \textbf{c}, \textbf{s}) = F_{\Theta}\Big(\gamma(\textbf{x}), \textbf{d}, W(\pi(\textbf{x}))\Big)
\end{equation}
The model is jointly trained on a collection of articulated objects belonging to the same category. Relative camera poses between multi-view images and the corresponding part segmentation labels are used for supervision.
\vspace{-0.5\baselineskip}

\subsection{Joint Attributes Estimation}
In this work, we consider 1D revolute joints. The joint attributes consist of the direction of the rotation axis $\mathbf{u}$ as well as a pivot point $\mathbf{v}$ on the rotation axis. Given an input image of the articulated object, we propose to infer the joint attributes from the predicted segmentation field through ray marching. For each pixel on the image plane, we cast a ray $\mathbf{r}(t) = \mathbf{o} + t\mathbf{d}$ starting from the camera center $\mathbf{o}$ along the direction $\mathbf{d}$ passing through that pixel. We then sample $K$ points $\{\mathbf{x}_k = \mathbf{r}(t_k)\}_{k=1}^K$ along the ray and feed them into the semantic NeRF to get their predicted density and part segmentation  $\{\sigma_k, \mathbf{s}_k\}_{k=1}^K$. Since the 1D revolute joint lies at the intersection of two parts, we filter the sampled points to collect points that are close to the intersection:
\begin{equation}
\mathbf{X}_{\text{intersection}} = \{\mathbf{x}_k \,|\, \argmax(\mathbf{s}_k) \neq \argmax(\mathbf{s}_{k+1}) \wedge \sigma_k \geq H\}
\end{equation}
where $H$ is a predefined threshold to remove points with low density. 
After collecting $\mathbf{X}_{\text{intersection}}$ from all the pixels, we can perform linear regression on these 3D points to estimate both rotation axis $\mathbf{u}$ and the pivot point $\mathbf{v}$.
\vspace{-0.3\baselineskip}

\subsection{Articulation-aware Volume Rendering}
After predicting the part segmentation field and joint attributes $\mathbf{J}$, we discuss a modified volume rendering procedure that allows us to perform view synthesis at unseen articulated poses.
Given an input articulated pose $\mathbf{a}$ specified by users, we construct deformation matrices \{$\mathbf{D}^i(\mathbf{a}, \mathbf{J})\}_{i=1}^{P+1}$ that describe the rigid transformation between part $i$ and the root part. 
% The detailed equation of $\mathbf{D}$ can be found in the supplementary material. 
%
During volume rendering, we deform the rays with each part's deformation matrix $\mathbf{D}^i$ and collect the outputs for all ${P+1}$ parts:
\begin{equation}
    \{\sigma^i, \mathbf{c}^i, \mathbf{s}^i = F_\Theta\Big(\gamma(\mathbf{D}_i\mathbf{x}), \mathbf{d}, W(\pi(\mathbf{D}_i\mathbf{x}))\Big)\}_{i=1}^{P+1}
\end{equation}
To merge these outputs for articulation-aware volumetric rendering, we weight all the fields with predicted part segmentation $\hat{p}$, where $\hat{p}^i$ indicates the estimated probability of being classified as part $i$.
The predicted color $\hat{\mathbf{C}}(\mathbf{r})$ and segmentation $\hat{\mathbf{S}}(\mathbf{r})$ are therefore the weighted sum of each part:
\begin{equation}
    \hat{\mathbf{C}}(\mathbf{r}) = \sum_{k=1}^{K} \hat{T}(t_k) \sum_i^{P+1} \hat{p}^i(t_k) (1-\exp(-\sigma^i_k(t_{k+1}-t_k)))\mathbf{c}^i(t_k)
\end{equation}
\begin{equation}
    \hat{\mathbf{S}}(\mathbf{r}) = \sum_{k=1}^{K} \hat{T}(t_k) \sum_i^{P+1} \hat{p}^i(t_k) (1-\exp(-\sigma^i_k(t_{k+1}-t_k)))\mathbf{s}^i(t_k)
\end{equation}
where the accumulated transmittance is
\begin{equation}
\hat{T}_k = \text{exp} (-\sum_{k' < k} \sum_i^{P+1} \hat{p^i}(t_k) \sigma^i_{k'} (t_{k'+1} - t_{k'}))
\end{equation}

\begin{figure*}[t]
    \centering
    \includegraphics[width=0.98\textwidth]{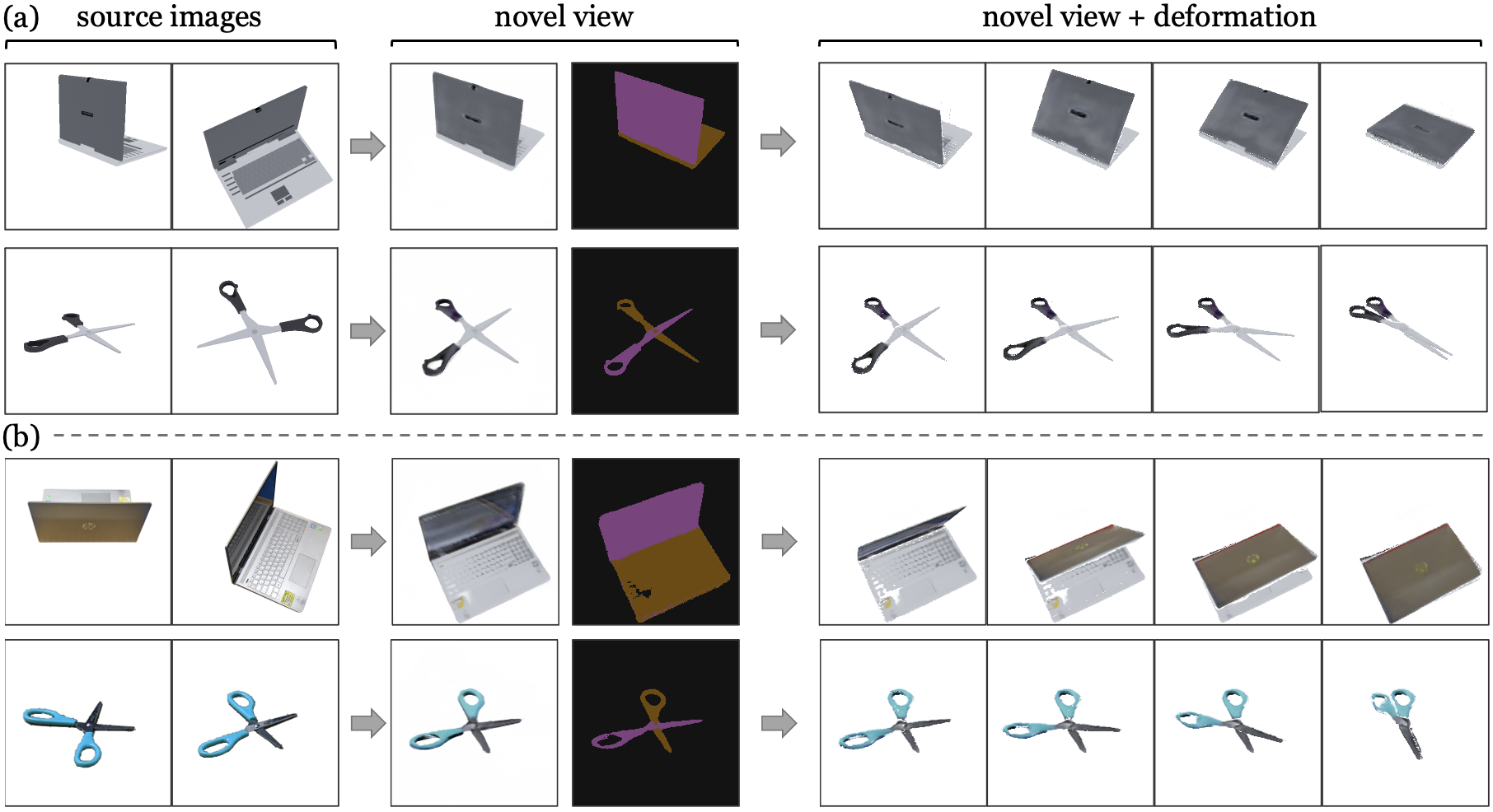}
    \vspace{-.5\baselineskip}
    \caption{The typical results on (a) synthetic data and (b) real-world data. We can find that the part in the object is consistently deformed with the joint parameter. 
    % See \url{https://bit.ly/3iWoque} for more qualitative results.
    }
    \label{fig:simu_qualitative}
    \label{fig:real_qualitative}
    \vspace{-0.5\baselineskip}
\end{figure*}

\subsection{Articulated Pose Estimation} \label{sec:ape}
Here we explain how we perform category-level articulated object pose estimation with CLA-NeRF. We assume the semantic NeRF $F_\Theta$ of an articulated object has already been predicted from source images and both the camera intrinsics and extrinsics are known. The goal is to estimate the articulated pose $\mathbf{a}$ of a given input image $I$. Unlike CLA-NeRF's training procedure which optimizes $\Theta$ using image observations and part segmentations, we instead solve the inverse problem~\cite{arxiv20_inerf} of recovering the articulated pose $\mathbf{a}$ given the weights $\Theta$ and the image $I$:
\begin{equation}
\hat{\mathbf{a}} = \text{argmin}_{\mathbf{a} \in \mathbf{A}}\mathcal{L}_{\text{color}}(\mathbf{a}|\Theta, \mathbf{d}, \mathbf{J})
\end{equation}
To solve this optimization problem, we iteratively perform gradient-based optimization to minimize the residuals between the rendered image and the observed image.

% \vspace{-0.5\baselineskip}

% \begin{figure*}[t]
%     \centering
%     % \includegraphics[width=1\textwidth]{figures/qual_hor.png} 
%     \includegraphics[width=1\textwidth]{figures/simple_qual.png}
%     % \vspace{-1.1\baselineskip}
%     \caption{The typical results on (a) synthetic data and (b) real-world data. We can find that the part in the object is consistently deformed with the joint parameter. See \url{https://bit.ly/3iWoque} for more qualitative results.}
%     \label{fig:simu_qualitative}
%     \label{fig:real_qualitative}
%     % \vspace{-0.5\baselineskip}
% \end{figure*}

\section{Experiments} \label{sec:experiment}
We evaluate CLA-NeRF on three different tasks: view synthesis, part segmentation, and articulated pose estimation.

% \vspace{-0.5\baselineskip}
% \subsection{Implementation Detail} \label{sec:implementation_detail}
% We plan to make our code and dataset publicly available, and the hyperparameters and detailed neural network architecture will be included.
% We present the hyperparameters, detailed neural network architecture, and computational complexity in the supplementary material.

\subsection{Dataset}
\paragraph{Synthetic data} We consider the ``laptop'', ``scissors'',  ``eyeglasses'', ``stapler'' and ``pliers'' classes of SAPIEN \cite{CVPR20_SAPIEN,arxiv15_shapenet,CVPR19_partnet} with 46, 54, 65, 24 and 23 instances respectively. We split these instances into two sets: training and held-out. We train on 200 observations of each training instance at a resolution of $200 \times 200$ pixels. Camera poses are randomly generated on a sphere with the object at the origin. Transparencies and specularities are disabled. We further render the held-out instances to construct a dataset for performance evaluation.

\paragraph{Real-world data}
To further test our method, we manually collect real-world images and the corresponding camera poses for the “laptop” and “scissors” with articulate poses at $[0^{\circ}, 30^{\circ}, 60^{\circ}, 90^{\circ}]$.

% \vspace{-0.5\baselineskip}
\subsection{View Synthesis and Part Segmentation} \label{sec:view}
\begin{table}[h]
    \setlength{\tabcolsep}{3.2pt}
    % \fontsize{8}{9.5}\selectfont
    \caption{Quantitative result including segmentation, articulate pose, and novel view synthesis of our framework evaluated on the dataset generated from SAPIEN\cite{CVPR20_SAPIEN}.}
    \centering
    \begin{tabular}{ |c|c c c c| c c| } 
        \hline
        \multicolumn{1}{|c|}{} &
        \multicolumn{4}{|c|}{Novel View Synthesis} &
        \multicolumn{2}{|c|}{Segmentation} \\
        \hline
         & MSE$\downarrow$ & PSNR$\uparrow$ & SSIM$\uparrow$ & LPIPS$\downarrow$ & Pixel Acc$\uparrow$ &  mIoU$\uparrow$ \\
        \thickhline
        Laptop & 0.0811 & 23.89 & 0.94 & 0.1323 & 0.981 & 0.971 \\ 
        Scissors & 0.0722 & 24.01 & 0.92 & 0.1456 & 0.989 & 0.969 \\ 
        Eyeglasses & 0.0991 & 23.72 & 0.89 & 0.1755 & 0.973 & 0.941 \\
        Stalper & 0.0771 & 26.91 & 0.96 & 0.1022 & 0.969 & 0.940 \\
        Pliers & 0.0413	& 25.90 & 0.96 & 0.0711 & 0.971 & 0.940 \\
        \hline
    \end{tabular}
    \label{table:simu_quantitative}
    \vspace{-0.5\baselineskip}
\end{table}

\begin{table}[h]
    \caption{Quantitative results for our real-world data.}
    \centering
    % \vspace{-1\baselineskip}
    \begin{tabular}{ |c|c c c c| } 
        \hline
        category & MSE$\downarrow$ & PSNR$\uparrow$ & SSIM$\uparrow$ & LPIPS$\downarrow$  \\ 
        \hline
        Laptop & 0.1021 & 22.12 & 0.93 & 0.1600 \\ 
        Scissors & 0.1281 & 23.39 & 0.92 & 0.1492 \\ 
        \hline
    \end{tabular}
    \label{table:real_quantitive}
    \vspace{-0.8\baselineskip}
\end{table}
We show the qualitative results in Fig.~\ref{fig:simu_qualitative} and quantitative results in Table~\ref{table:simu_quantitative} for the synthetic data. We find that CLA-NeRF successfully renders the held-out object at different articulated poses. 
For the real data, we report the quantitative results in Table \ref{table:real_quantitive} and qualitative results in Fig. \ref{fig:real_qualitative}. The used metrics are MSE/PSNR/SSIM (higher is better) and LPIPS \cite{zhang2018perceptual} (lower is better). We found that the network trained on synthetic data effectively infers the shape and texture of the real object, suggesting our model can transfer beyond the synthetic domain. 
% \vspace{-0.5\baselineskip}

% \begin{figure*}
%     \vspace{-1\baselineskip}
%     \centering
%     \includegraphics[width=1\textwidth]{figures/articulate.png}
%     \vspace{-1.5\baselineskip}
%     \caption{The qualitative results for articulated pose estimation. 
%     Rendered images based on the estimated articulated pose at optimizing steps $n$ and the target observation. Our method is able to recover the articulated pose by aligning the images.}
%     \vspace{-1.\baselineskip}
%     \label{fig:articulate}
% \end{figure*}

\subsection{Articulated Pose Estimation} \label{sec:exp_ape}
\begin{table*}[ht]
    \caption{Quantitative results for articulated pose estimation and joint localization. We show the pose error $\mathbf{a}_{\text{error}}$, angle error $\mathbf{u}_{\text{error}}$, distance error $\mathbf{v}_{\text{error}}$. For the real-world results, please see Sec. \ref{sec:exp_ape}.}
    \centering
    \begin{tabular}{c|c c c | c c c | c c c | c c c} 
        \hline
        \multicolumn{1}{c|}{Dataset} &
        \multicolumn{9}{|c|}{Synthetic} &
        \multicolumn{3}{|c}{Real-World} \\
        \hline
        \multicolumn{1}{c|}{Apporach} &
        \multicolumn{3}{|c}{Ours} &  
        \multicolumn{3}{|c}{ScrewNet~\cite{arxiv20_screwnet}} &  
        \multicolumn{3}{|c}{Abbatematteo et al. \cite{corl20_learn}} &  
        \multicolumn{3}{|c}{Ours} \\ 
        \hline
         & $\mathbf{a}_{\text{error}}$ & $\mathbf{u}_{\text{error}}$ & $\mathbf{v}_{\text{error}}$ & $\mathbf{a}_{\text{error}}$ & $\mathbf{u}_{\text{error}}$ & $\mathbf{v}_{\text{error}}$ & $\mathbf{a}_{\text{error}}$ & $\mathbf{u}_{\text{error}}$ & $\mathbf{v}_{\text{error}}$ & $\mathbf{a}_{\text{error}}$ (sim2real)& $\mathbf{a}_{\text{error}} $ (generalize)& $\mathbf{a}_{\text{error}}$(overfit) \\ 
        \thickhline
        Laptop & 0.138 & 0.010 & 0.091 & 0.129 & 0.019 & 0.062 & 0.137 & 0.012 & 0.041 & 0.179 & 0.174 & 0.179 \\ 
        Scissors & 0.130 & 0.016 & 0.120 & 0.116 & 0.149 & 0.136 & 0.131 & 0.037 & 0.041 & 0.179 & 0.170 & 0.170  \\ 
        Eyeglasses & 0.151 & 0.109 & 0.071 & 0.141 & 0.140 & 0.136  & 0.149 & 0.108 & 0.082 & - & - & -  \\
        Stalper & 0.182 & 0.021 & 0.010 & 0.119 & 0.146 & 0.101  & 0.172 & 0.031 & 0.008 & - & - & -  \\
        Pliers & 0.171 & 0.010 & 0.010 & 0.121 & 0.132 & 0.102  & 0.183 & 0.009 & 0.009 & - & - & -  \\
        \hline
    \end{tabular}
    \label{table:real_articulate}
    \vspace{-1\baselineskip}
\end{table*}

\begin{figure}
    % \vspace{-1.5\baselineskip}
    \begin{center}
        \includegraphics[width=0.45\textwidth]{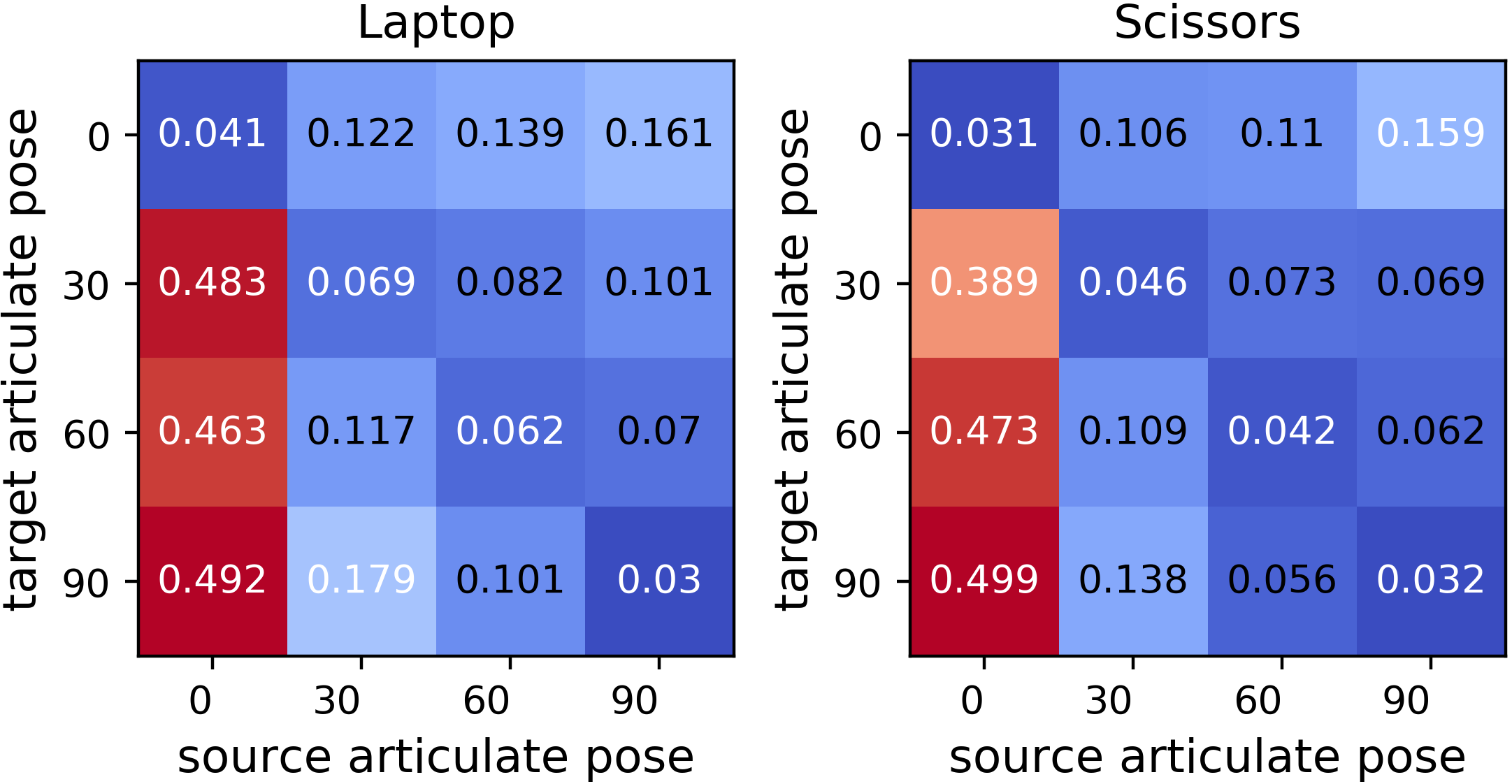}
    \end{center}
    \vspace{-0.3\baselineskip}
    \caption{Error heatmap of articulate pose estimation.}
    \label{fig:articulate_matrix}
    \vspace{-0.7\baselineskip}
\end{figure}

\begin{table}[ht]
    \caption{We also evaluate our approach with pose error $\mathbf{a}_{\text{error}}$, angle error $\mathbf{u}_{\text{error}}$, distance error $\mathbf{v}_{\text{error}}$ on Shape2Motion validation set. Note that ANCSH~\cite{cvpr20_ANCSH} requires depth to estimate pose.}
    \label{table:ancsh}
    \centering
    \begin{tabular}{|c|c c c | c c c|} 
        \hline
        \multicolumn{1}{|c|}{Approach} &
        \multicolumn{3}{|c|}{Ours} &
        \multicolumn{3}{|c|}{ANCSH~\cite{cvpr20_ANCSH}} \\
        \hline
         &  $\mathbf{a}_{\text{error}}$ & $\mathbf{u}_{\text{error}}$ & $\mathbf{v}_{\text{error}}$ &   $\mathbf{a}_{\text{error}}$ & $\mathbf{u}_{\text{error}}$ & $\mathbf{v}_{\text{error}}$ \\ 
        \thickhline
        Laptop & 0.179 & 0.011 & 0.110 & 0.169  & 0.009 & 0.017  \\ 
        Eyeglasses &  0.169 & 0.109 & 0.091 & 0.076  & 0.039 & 0.016  \\
        \hline
    \end{tabular}
    \vspace{-1\baselineskip}
\end{table}
The results on synthetic data and real-world data are presented in Table \ref{table:real_articulate}. Fig. \ref{fig:articulate_matrix} shows the L1 error with different articulate poses of source images and target images on the real-world dataset. We can first find that the error is the lowest when no deformation is required. 
Second, if the articulated poses are too different, the estimation will be less accurate. It is because we only optimize articulated pose with $\mathcal{L}_{\text{color}}$ and local minimum occur during the optimization process.

To test the limit of our method, we compare our method with ScrewNet~\cite{arxiv20_screwnet} and Abbatematteo et al. \cite{corl20_learn} on our dataset (Table \ref{table:real_articulate}).
Besides, we also evaluate CLA-NeRF on the Shape2Motion dataset without fine-tuning. The results compare against ANSCH~\cite{cvpr20_ANCSH} are shown in Table~\ref{table:ancsh}. Despite not using depth images as inputs and not finetuned, we find our model to only perform slightly worse than ANSCH~\cite{cvpr20_ANCSH} and ScrewNet~\cite{arxiv20_screwnet} and \cite{corl20_learn}. It shows that the proposed representation is a promising direction for category-level articulated pose estimation.

To understand whether the articulated pose estimation can be improved, we finetune the model in two manners. First, we finetune the framework with a set of real-world objects, then, we test it on unseen real-world objects. Second, we directly finetune the framework on specific object and test on it. These fine-tuning approaches are labeled as \textbf{generalize} and \textbf{overfit} in Table \ref{table:real_articulate}, respectively. Only minor improvement is observed.

\subsection{Failure Cases} \label{sec:failure}
\begin{figure}
    % \vspace{-1.2\baselineskip}
    \begin{center}
    \includegraphics[width=0.48\textwidth]{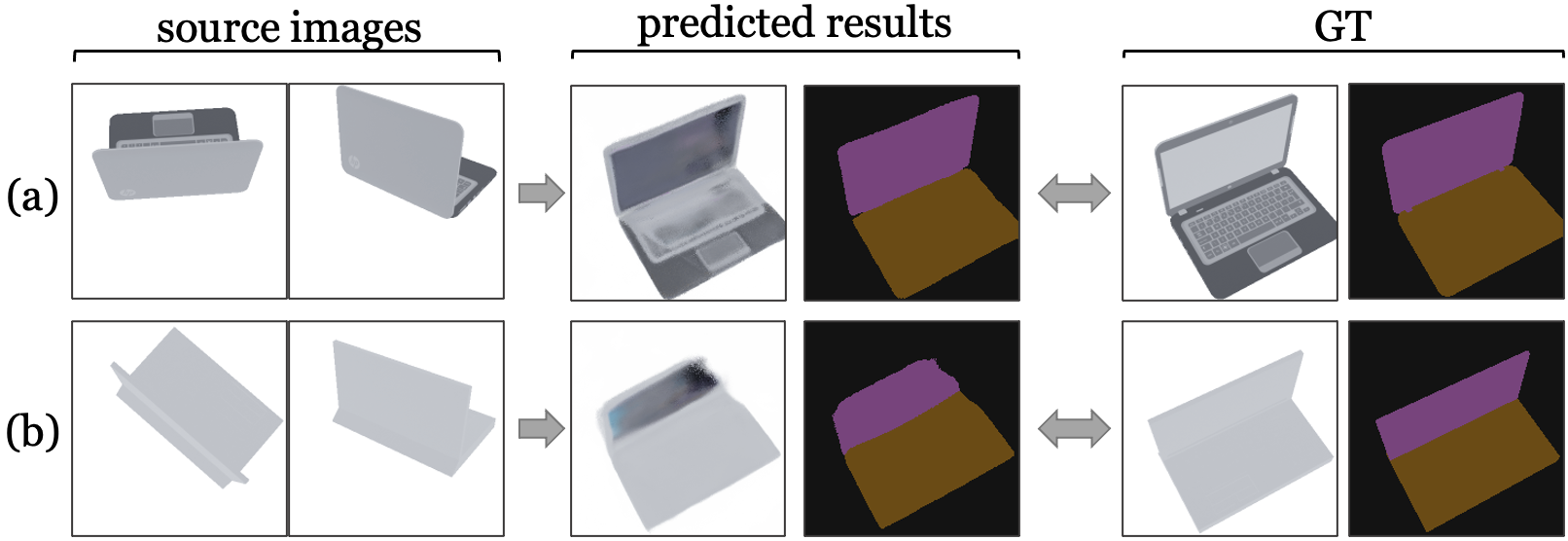}
    \end{center}
    \vspace{-.2\baselineskip}
    \caption{Failure cases. (a) Incorrect keyboard appearance due to missing observation in source images. (b) Incorrect geometry due to lack of texture.}
    \label{fig:failure}
    \vspace{-.2\baselineskip}
\end{figure}

Despite the promising results shown in Sec. \ref{sec:view} and Sec. \ref{sec:exp_ape}, there are some failure cases that need to be discussed. First, since our framework only takes few instances as conditions, if query camera poses are highly distinct from the source camera pose, the appearance may have some defect. From Fig. \ref{fig:failure} (a), the predicted screen color is not the same as the ground truth. However, the geometry of the object and its part segmentation are reconstructed correctly, so the articulated pose estimation isn't affected by this issue. Besides, despite the color of the screen isn't accurate, the appearance still matches the normal appearance of the laptop screen.
Second, for the CAD model without texture, it is hard for our framework to correctly infer its geometry. From Fig. \ref{fig:failure} (b), we find that the shape of the screen is distorted. Besides, our joint localization is inaccurate if the source images are closed laptop. It is because the NeRF model the two part of laptop attach with each other, so the intersection points form a surface. Therefore, the joint axis can't be estimated correctly.

\section{Conclusion} \label{sec:conclusion}
We propose a framework that only takes few instances of an articulate object with different viewpoints as references; then, infers the corresponding deformable neural radiance field to predict the image and part segmentation with the specified camera pose. With the well-trained framework, the articulate pose of an object can be estimated via inversely optimize the deformation condition. In the experiments, we evaluate the framework in both synthesis objects collected from SAPIEN and our manually collected real-world data. In all cases, our method shows realistic deformation results and accurate articulated pose estimation.
\section{Acknowledge}

This project is funded by Ministry of Science and Technology of Taiwan (MOST 109-2634-F-007-016) and supported by NOVATEK Fellowship.

\bibliographystyle{unsrt}
\bibliography{reference}
\addtolength{\textheight}{-12cm}   % This command serves to balance the column lengths



% \section*{ACKNOWLEDGMENT}

% The preferred spelling of the word ÒacknowledgmentÓ in America is without an ÒeÓ after the ÒgÓ. Avoid the stilted expression, ÒOne of us (R. B. G.) thanks . . .Ó  Instead, try ÒR. B. G. thanksÓ. Put sponsor acknowledgments in the unnumbered footnote on the first page.

%%%%%%%%%%%%%%%%%%%%%%%%%%%%%%%%%%%%%%%%%%%%%%%%%%%%%%%%%%%%%%%%%%%%%%%%%%%%%%%%

% References are important to the reader; therefore, each citation must be complete and correct. If at all possible, references should be commonly available publications.
%\bibliographystyle{unsrt}
%\bibliographystyle{IEEEtran}
% \bibliography{reference}
%\bibliography{reference}  % .bib
% \begin{thebibliography}{99}

% \end{thebibliography}
\end{document}